\pdfoutput=1
\documentclass[11pt]{article}

\usepackage[final]{acl}

\usepackage{times}
\usepackage{latexsym}

\usepackage[T1]{fontenc}
\usepackage[utf8]{inputenc}

\usepackage{microtype}

\usepackage{graphicx}

\usepackage{inconsolata}

\title{Comparing Two Model Designs for Clinical Note Generation; Is an LLM a Useful Evaluator of Consistency? }

\author{Nathan Brake \and Thomas Schaaf\\
  Solventum, Health Information Systems\footnotemark[1]  \\
  \texttt{nbrake,tschaaf@solventum.com} \\}

\begin{document}
\maketitle
\footnotetext[1]{Solventum is a standalone healthcare technology company created following a spin-off of 3M's healthcare division. }
\begin{abstract}

 Following an interaction with a patient, physicians are responsible for the submission of clinical documentation, often organized as a SOAP note. A clinical note is not simply a summary of the conversation but requires the use of appropriate medical terminology. The relevant information can then be extracted and organized according to the structure of the SOAP note. In this paper we analyze two different approaches to generate the different sections of a SOAP note based on the audio recording of the conversation, and specifically examine them in terms of note consistency. The first approach generates the sections independently, while the second method generates them all together. In this work we make use of PEGASUS-X Transformer models and observe that both methods lead to similar ROUGE values (less than 1\% difference) and have no difference in terms of the Factuality metric. We perform a human evaluation to measure aspects of consistency and demonstrate that LLMs like Llama2 can be used to perform the same tasks with roughly the same agreement as the human annotators. Between the Llama2 analysis and the human reviewers we observe a Cohen Kappa inter-rater reliability of 0.79, 1.00, and 0.32 for consistency of age, gender, and body part injury, respectively. With this we demonstrate the usefulness of leveraging an LLM to measure quality indicators that can be identified by humans but are not currently captured by automatic metrics. This allows scaling evaluation to larger data sets, and we find that clinical note consistency improves by generating each new section conditioned on the output of all previously generated sections.

\end{abstract}

\section{Introduction}

As a part of a physician's workload, the Electronic Health Record (EHR) has become an important tool for  documenting patient information that is used for care and billing purposes. A SOAP note is a common framework for structuring a record of a Doctor Patient Conversation (DoPaCo) that consists of Subjective, Objective, Assessment, and Plan sections. 

Two common components inside of the Subjective section are the "Chief Complaint" (CC) and "History of Present Illness" (HPI) sections. Chief Complaint is normally a brief one sentence statement about the reason for the patient's visit to the physician. For example: "Patient presents for evaluation of left foot pain". This information is mentioned in the DoPaCo, but may also be provided as a part of the patient's admission documentation. History of Present Illness is usually a multi-sentence or paragraph description of relevant patient information that was discussed in the DoPaCo. For example, it may contain snippets such as:  "Patient is a 60-year-old male who reports left foot pain after having his foot run over by a tractor in 2021 . . . He had surgery on his foot in 2022 and is still experiencing pain . . . The patient reports a history of osteoporosis."

The Assessment section is normally a brief description or list of the doctor's assessment, e.g. "Assessment: Left Foot Pain" that may be mentioned during the DoPaCo or directly entered or dictated into an EHR system following the encounter. The Plan section is a description of a path forward and commonly has a more narrative style such as: "Plan: I have personally reviewed the findings with the patient today. He is scheduled for a left total knee arthroplasty soon. I anticipate that this procedure will also help with his left foot pain. At this time, we will hold off on a new orthotic prescription. He will follow up with me at the end of February." For our experiment we combine Assessment and Plan to be considered as a single section.

In this work we present clinical notes that contain the Chief Complaint (CC), History of Present Illness (HPI), and Assessment \& Plan (A\&P) sections. We omit the Objective section because at the time of writing, the Objective section commonly contains information from a physical examination that has not been directly verbalized during the DoPaCos that are used in our dataset. We leave to future work the exploration of generating the Objective section of a SOAP note using a DoPaCo and incorporating additional information that was not verbalized during the encounter.

In order to ease the burden on a physician and accelerate workflows, recent research in Natural Language Processing (NLP) techniques are being explored to automatically generate SOAP notes using variants of the Transformer architecture \cite{NIPS2017_3f5ee243}.  A common approach to the automatic generation of a SOAP note is to use an automatic speech recognition (ASR) system to create a transcript of the DoPaCo, based upon an audio recording of the encounter.  

In this work, we present a comparison of two designs for generating a SOAP note. From the pre-trained PEGASUS-X model \cite{phang2022investigating}  , we train a single fine-tuned model (GENMOD) to produce a clinical note, as well as 3 individual fine-tuned models (SPECMOD) to each produce a single section.

In order to compare these two note generation designs,  ROUGE \cite{lin-2004-rouge} is a common automatic metric for measuring the performance of summarization models. However, it is not always a reliable proxy for human preference, and a model with a lower ROUGE score may be preferred by humans \cite{ziegler2020finetuning}. In particular, for comparing the two model approaches we expect that the SPECMOD design would be more likely than GENMOD to have conflicting content between sections, since in SPECMOD the output of a section such as A\&P is not conditioned on the output text of any other section, e.g. HPI. For example, if the DoPaCo does not make clear reference to the gender of the patient, we expect that SPECMOD would be more likely than GENMOD to refer to the patient as female in HPI and mistakenly refer to the patient as a male in A\&P.  In this work we present an approach to use the Llama2 LLM \cite{touvron2023Llama}  as an additional measure of model quality for criteria not clearly captured by existing automatic metrics.

\section{Related Work}

Clinical note generation can be viewed as a summarization problem, since it can involve the use of a DoPaCo as input to summarize the content into a document that uses the appropriate clinical terminology and style. Recent discoveries in deep learning based NLP have enabled advancements in the creation of clinical notes from DoPaCos \cite{krishna-etal-2021-generating, su22b_interspeech, zhang-etal-2021-leveraging-pretrained, michalopoulos-etal-2022-medicalsum} . \cite{zhang-etal-2021-leveraging-pretrained} investigate the use of a fine-tuned BART model \cite{lewis2019bart} for generation of the HPI section of a clinical note. Similarly, \cite{singh-etal-2023-large} generate a clinical note that contains an HPI, A\&P, and Physical Examination section, and train a separate model for each section.  \cite{ID197_Research_Paper_2023} seeks to improve the consistency of SOAP notes through the integration of section tokens and section-specific cross-attention parameters to encoder-decoder models. This approach uses a single BART model with a modified cross-attention mechanism to generate a SOAP note based on extracted segments of the DoPaCo, and produces a single section at a time based upon the special token pre-pended to the input conversation.

In order to evaluate the quality of these generated clinical notes, it can be difficult to find a reliable proxy for human evaluation. With the advent of ChatGPT/GPT-4 \cite{openai2023gpt4} and Llama2 \cite{touvron2023Llama}, the use of LLMs as a proxy for human evaluation is a popular subject of recent research \cite{zeng2023evaluating, chiang2023large, Gilardi_2023, zheng2023judging}. \cite{chiang2023large} found LLM based evaluation results to be consistent with human evaluation across several NLP tasks in terms of cohesiveness, among other criteria. \cite{zheng2023judging} finds that GPT-4 has the ability to match human preferences when comparing two different LLM generated answers to a question. \cite{liu2023geval} presents a framework called G-Eval which uses chain-of-thought prompting and form-filling to evaluate text outputs for coherence. The LLM Prometheus  \cite{kim2023prometheus} is a fine-tuned Llama2-13B LLM designed to act as an evaluator LLM that is aligned with human preferences. \cite{xie2023enhancing} investigate the use of GPT-4 to evaluate medical notes in terms of factuality. To the best of our knowledge, this work is the first to evaluate the performance of an LLM in reviewing clinical notes for specific criteria of consistency. 

ACI-Bench \cite{yim2023acibench} provides results from comparison of full-note vs division-based note generation techniques, which corresponds to our GENMOD and SPECMOD designs, respectively. Their work finds that SPECMOD results in higher performance then GENMOD. However, ACI-Bench uses a smaller dataset of 207 encounters with an average conversation length of ~1,300 tokens. Using our larger proprietary dataset with more average tokens per conversation, we seek to understand whether the comparison of GENMOD to SPECMOD is sensitive to the size of the dataset as well as the dimension upon which the two models are measured (i.e. in terms of full note consistency).

\section{Data}

The DoPaCos in the dataset for this experiment come from an asynchronous scribing configuration, similar to that which is described in detail in \cite{9688118}. The physician is aware that the audio is being provided to a human scribe after the encounter, and as a result, the audio may contain dictated portions where the doctor directly instructs the scribe regarding information that should be included in the note. The physicians in the dataset represent a mix of 20 different specialties, identified using the National Provider Identification (NPI) Registry (https://npiregistry.cms.hhs.gov)

We create three dataset splits for the experiment: Training, Validation, and Test. Information about these splits can be found in Table~\ref{tab:dataset}. The training, validation, and test data have similar length distributions for the input conversations and target clinical notes.

\begin{table*}
\centering
\begin{tabular}{lccccll}
\hline
Split & Rows & Spec & Phys. & Avg Dur (min) & Avg Tok. in Conv. &Avg Tok in Clinical Note\\
\hline
Train & 9800 & 20  & 82 & 16.58  & 3,312& 524\\
Val & 516 & 15 & 54 & 16.14  & 3,430 & 525\\
Test & 543 & 14 & 50 & 17.21  & 3,237& 513\\\hline
\end{tabular}
\caption{Dataset Information, including the number of different Specialties (Spec),  Physicians (Phys.), and the average Dopaco Duration (Avg Dur)}
\label{tab:dataset}
\end{table*}

The MediQA \cite{ben-abacha-etal-2023-overview} task uses the ACI-Bench \cite{yim2023acibench} dataset for clinical note summarization, which contains 207 encounters of average length ~1,300 tokens. This precludes it from being substituted as a training dataset for this experiment due to the small dataset size and shorter average token length of the conversations. Further, 112 of the 207 ACI-Bench encounters come from the ambient clinical intelligence setting, which does not match the asynchronous scribing domain that our experiment addresses.

\section{Methodology}

\subsection{Training}
For this experiment, we train models to generate clinical notes that contain 3 sections: CC, HPI, and A\&P.

We train the transformer based PEGASUS-X large model \cite{phang2022investigating}, which is a 568M parameter model that builds on top of the PEGASUS model \cite{pmlr-v119-zhang20ae} and uses a modified attention mechanism in order to support longer input sequences of up to 16,384 tokens, and has been additionally pre-trained using long sequence data. We expect our method to extend to other transformer based models as well. Training is performed on a compute cluster of 8 NVIDIA A100 GPUs.

\begin{figure}[htp]
    \centering
    \includegraphics[width=7cm]{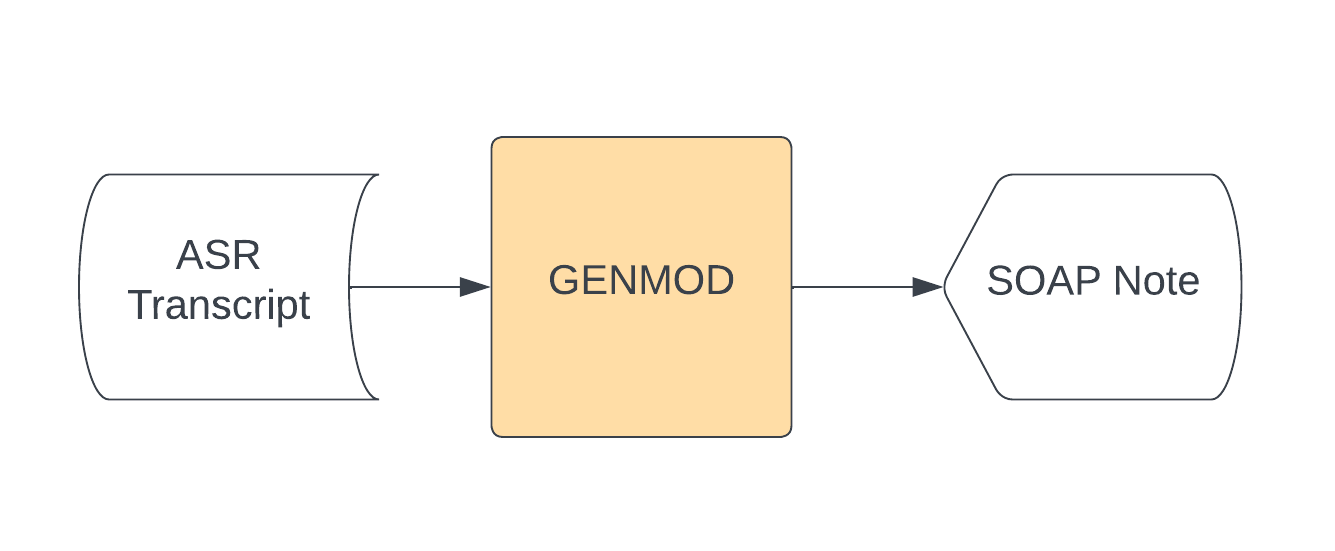}
    \caption{Creation of SOAP note with GENMOD}
    \label{fig:genmod}
\end{figure}

\begin{figure}[htp]
    \centering
    \includegraphics[width=7cm]{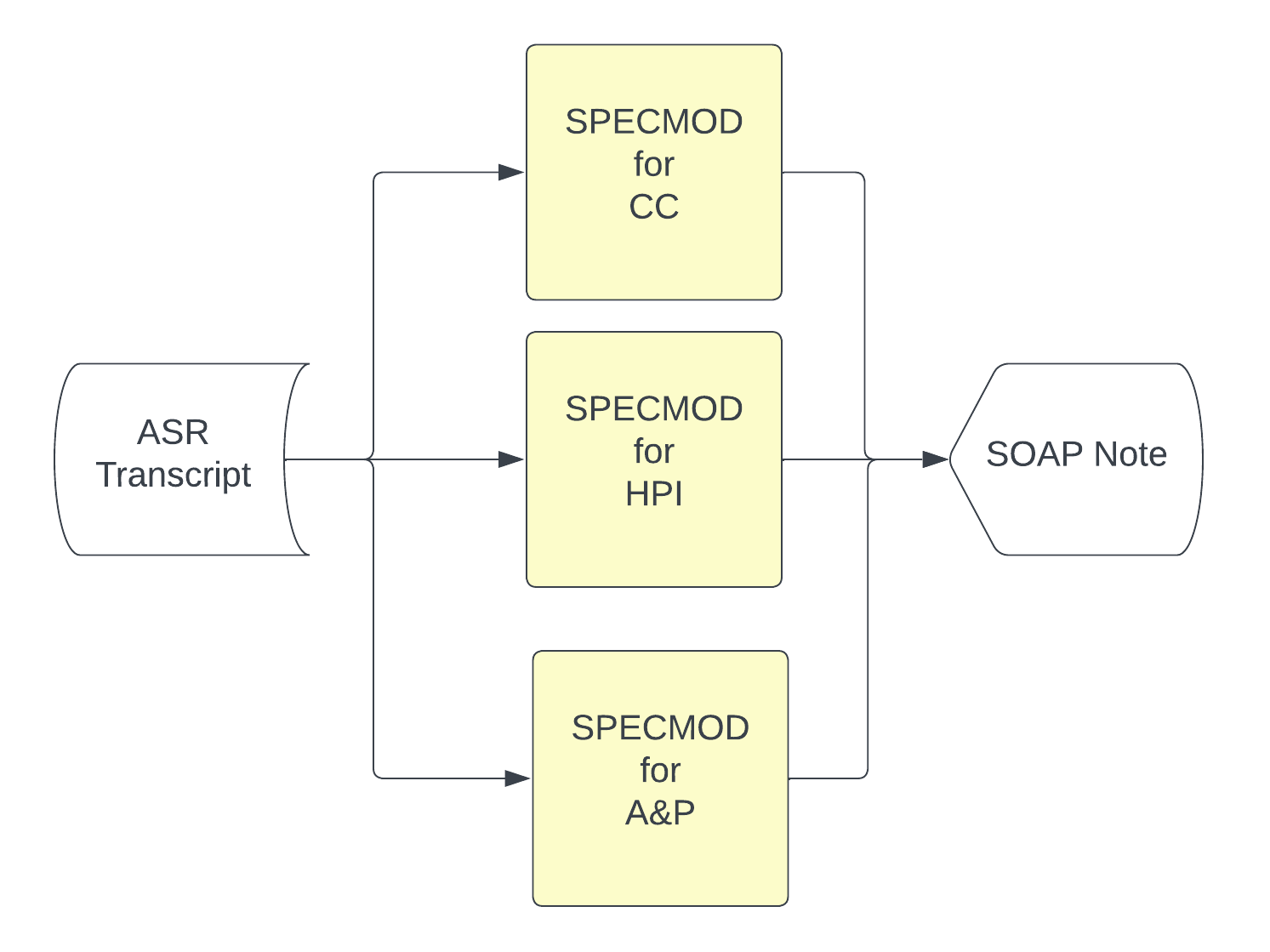}
    \caption{Creation of SOAP note with SPECMOD}
    \label{fig:specmod}
\end{figure}

Two methods for clinical note generation are trained:
\begin{enumerate}
    \item General Model (GENMOD). Shown in Figure \ref{fig:genmod}, this is a single PEGASUS-X model that produces the entire note in a single generation step. The model is trained to output 6 added special tokens, one token to indicate the start of a section (e.g. "<history\_of\_present\_illness>"), and one to denote the end of a section (e.g. "</history\_of\_present\_illness>"). This design allows for easy parsing of the note when scoring each section individually. It also allows for scoring of the note as a whole, since the special tokens can be skipped during the tokenizer decoding step. When generating automatic metrics (ROUGE, Factuality) as well as during human review, these special tags are omitted so as not to bias the model output to a higher score. Otherwise, the special tags would always be present in both the generated and reference note and would artificially inflate metrics that check for n-gram overlaps, such as ROUGE.
    \item Specialized Model (SPECMOD). Shown in Figure \ref{fig:specmod}, these are 3 individual PEGASUS-X models, one model trained for each section. Each model is trained to output a single section. For evaluations that use a fully generated clinical note, the transcript is provided to each model which generates its individual section. These outputs are then combined to form the full note. 
\end{enumerate}

\subsection{Evaluation}
\label{sec:eval}
After training each model configuration, we evaluate the model generated notes using common automatic metrics, human reviewers, and the Llama2 LLM.  The notes are generated from the models using beam search generation with a beam size of 4, and a No Repeat N-Gram Size (NRNS) of 5. The automatic metrics indicate the performance of the models using the metrics commonly used to evaluate a model. The human and Llama2 evaluations help to expose differences between GENMOD and SPECMOD generated notes that are not captured by ROUGE or Factuality.

The automatic metrics used for evaluation are ROUGE \cite{lin-2004-rouge} and Factuality \cite{glover-etal-2022-revisiting}. ROUGE is a common metric for use in measuring the performance of abstractive summarization tasks, and Factuality is a metric for measuring the factual content of the output compared to ground truth. The Factuality metric has been trained for medical concepts.

For the human and LLM evaluation, we collect data for four different criteria related to consistency, for a subset of 40 generated clinical notes found in the dataset test split. The cost and time required of human reviewers prevents the evaluation of the entire test split for this experiment. The humans and LLM are not provided with the reference (ground truth note) or the conversation transcript. They are displayed only a single generated note and asked to evaluate the following items:

\begin{enumerate}
    \item Age consistency. The clinical notes may state the age of the patient in multiple sections of the note, and the reviewers are tasked with identifying whether that age is consistent throughout. For example, if the CC section mentions that the patient is 65 years old, but the HPI states that the patient is 70 years old, this is marked as inconsistent. If the age is only mentioned in a single location, the note is marked as consistent. 
    \item Gender consistency. Similar to age consistency, the clinical note should refer to the patient by the same gender throughout the clinical note. For example, if the HPI section refers to the patient as a female and uses she/her/hers pronouns, the A\&P should also use she/her/hers pronouns.
    \item Body Part consistency. In some cases (most commonly for orthopedic specialty), the patient is being seen for a specific injury, whether that is a left leg, a right hip, etc. If a specific body part and location is mentioned, this should remain consistent throughout the note. If the CC and HPI discuss the right foot, but the A\&P addresses only the left foot or a right hip, this is inconsistent in terms of body part.
    \item Coherence. The CC/HPI/A\&P should be coherent with each other. The content of the A\&P should be reasonable based on the HPI and CC sections. There can be some additional content in the A\&P but it should not be contradictory to the HPI. For example, if the visit is for a followup, the A\&P section may contain much information that is not stated in the HPI. As long as the content of the A\&P looks reasonable given the HPI, that is considered coherent. However, if the HPI and CC are discussing depression, the A\&P should have something related to depression. If the HPI and CC are about depression but the A\&P is instead addressing diabetes, that is not consistent. Similarly, if the CC says the patient reports nausea but then the HPI says the patient is not experiencing nausea, that is not consistent.
\end{enumerate}

In a realistic production environment, information about the patient such as age and gender is expected to be available. However, since this information is not available in the experimental setting, we find that these items serve as a helpful proxy for evaluating how well systems like an LLM are able to identify specific issues.

\subsection{Human Evaluation}

The human evaluation is performed using 5 human reviewers. 3 reviewers are medical experts, while the remaining 2 reviewers are not medical experts. 

Each reviewer is shown the generated clinical note from GENMOD and SPECMOD for 40 different DoPaCo encounters from the test split, for a total of 80 notes. The clinical notes are presented to the reviewer one at a time in a randomized order. For each note they are asked to select whether each of the 4 evaluation criteria described in Section \ref{sec:eval} are satisfied. Because they are not provided the ground truth clinical note, they are not evaluating whether the content is correct when measured against a reference note, but only that the note satisfies the specified criteria in terms of consistency with itself.

\subsection{LLM Evaluation}

The LLM used for evaluation is the Llama2 Chat model. The Llama2 family of models \cite{touvron2023Llama} are open source, trained by Meta AI, and provide close to state-of-the-art performance on a variety of tasks. We experiment with a variety of sizes: 7B, 13B, and 70B. For the 13B and 7B model we also experiment with quantization of the weights to 8 bits. For clarity we present only the results of the 70B full precision model, but include a discussion of observed changes in quality in Appendix \ref{sec:size}. We evaluate the model using zero-shot, 1-shot, 2-shot, and 3-shot in-context prompting, to maximize possible performance improvements that can be gained through in-context learning \cite{brown2020language}. The purpose of using the LLM as an evaluator is to construct a prompt for the LLM in such a way that the LLM ranking of the clinical notes will closely align with the human ranking, so that the LLM can be useful as a proxy for human evaluation.

The Llama2 model is provided the individual generated clinical note from GENMOD and SPECMOD from the same 40 DoPaCo encounters that were provided to the human reviewers. For each of the 80 notes, Llama2 is provided 4 separate prompts, 1 for each evaluation criteria. The system prompts are specific to each evaluation criteria and are provided in Appendix \ref{sec:sys_prompts}. In the zero-shot configuration, each evaluation request uses a new context window, such that its output is not being conditioned on any previous responses of the model. Each prompt requests that the model review the provided note and evaluate whether it meets a single criteria item (age/gender/body part consistency/coherent). In the 1-shot configuration, the model is provided a single example note and an accompanying answer and explanation. The example note is designed to be similar in style to a real note used for review, and is marked as consistent in the evaluation criteria being requested. In the 2-shot and 3-shot configuration, the model is provided at least one example of an inconsistent and one example of a consistent note in the evaluation criteria being requested, along with a detailed explanation about why that note met or did not meet the criteria. The example notes and answers were tailored to address corner cases that were observed when developing prompts on the evaluation dataset split. For example, an example note provided for the criteria of age specifically addresses how to handle the case of the patient age only being mentioned once throughout the whole note, even when other dates were mentioned: i.e. that the model should not try to infer age from random dates mentioned in the note (like dates of surgical procedures), but should only pay attention to specific references to the patient's age.

Similar to the human evaluation, the LLM is never provided the ground truth clinical note to compare against. It is provided a single hypothesis clinical note, example prompts when in the 1/2/3-shot configuration, and a single evaluation criteria request.

The prompts used with Llama2 were developed using notes generated from the evaluation set, and the final results reported are from the 80 notes generated by GENMOD and SPECMOD from the test set.

\begin{table*}
\centering
\begin{tabular}{llllllll}
\hline
Model& NRNS & Section& R-1 & R-2 & R-3  & R-L  & Fact \\ \hline
GENMOD   & 5  &CC& \textbf{80.9}& \textbf{72.0}& 64.1& \textbf{80.7}& \textbf{74.5}\\
SPECMOD  & 5   &CC& 79.1& 71.6& \textbf{64.3}& 78.9& 73.4\\ \hline
 GENMOD   & 5 & HPI& 54.2& 32.6& 23.6& 41.1&52.9\\
 SPECMOD & 5   & HPI& 54.2& \textbf{32.8}& \textbf{24.1}& \textbf{41.2}&\textbf{53.0}\\ \hline
 GENMOD & 5   & A\&P& 50.2& 36.1& 30.0& 42.3&69.4\\
 SPECMOD & 5   & A\&P& \textbf{52.4}& 3\textbf{8.8}& \textbf{32.8}& \textbf{45.3} &\textbf{72.2}\\ \hline
 GENMOD  & 5  & Full Note& 58.5& 37.1& 28.5& 43.1& 63.6\\
 SPECMOD & 5   & Full Note& \textbf{59.5}& \textbf{38.7} & \textbf{30.4} & \textbf{45.1} &\textbf{64.7}\\ \hline
GENMOD  & 12  & Full Note& 59.5 & 38.7 & 30.4 & 45.1 & 64.7 \\
 SPECMOD & 12 & Full Note& \textbf{59.7}& \textbf{39.1} & \textbf{30.9} & \textbf{45.4} & 64.7\\ \hline
\end{tabular}
\caption{Automatic Metrics of models scored against the test split, including their No Repeat N-gram Size (NRNS) used during generation}
\label{tab:automatic}
\end{table*} 
\section{Results and Discussion}

\subsection{Automatic Metrics}
\label{sec:auto_metrics}
The automatic metric results are reported in Table \ref{tab:automatic}. We include ROUGE-1 (R-1), ROUGE-2 (R-2), ROUGE-3 (R-3), ROUGE-L (R-L), and Factuality (Fact). 

During generation, a parameter No Repeat N-gram Size (NRNS) controls the maximum token sequence that can be repeated in a generation, and is a technique to reduce the repetition of an output. For the human evaluation and individual section automatic scoring, a NRNS of 5 is used. When using NRNS of 5, SPECMOD generally outperforms the GENMOD architecture across all sections by a score of less than 3\%. 

However, when relaxing the NRNS to 12, the gap between SPECMOD and GENMOD shrinks to a difference of less than 0.5\% ROUGE, and reaches the same Factuality score. The improvement of GENMOD scores when adjusting NRNS can be explained by the nature of the GENMODs generation design. Since GENMOD is producing the entire note in a single generation step, a low NRNS prevents content that was in earlier sections from re-appearing in later sections, even when that repetition is reasonable. For example, the phrase "patient has fractured left fibula" may be reasonable to appear in both the HPI and the A\&P for a clinical note, but would not have appeared when using an NRNS of 5 or lower. 

Human evaluation and LLM evaluation was performed only on notes generated using NRNS of 5 due to resource limitations, and we leave to future work the continued exploration of how modified generation and decoding strategies effect the quality and consistency of clinical notes.

\subsection{Human and LLM Review}

\begin{table*}[]
	\centering
	\begin{tabular}{cccccc}
		\hline
		\textbf{Group} & \textbf{Model} & \textbf{Age} & \textbf{Gender} & \textbf{BP} & \textbf{Coh}\\ \hline
		 Med experts & GENMOD & \textbf{99.17\%} & \textbf{98.33\%} & \textbf{99.17\%} & 70.83\% \\ 
		 Med experts & SPECMOD & 95.00\% & 95.83\% & 90.00\% & \textbf{75.00\%} \\ 
		\hline
		 Non-Med experts & GENMOD & \textbf{100.00\%} & \textbf{97.50\%} & \textbf{92.50\%} & \textbf{63.75\%} \\ 
		 Non-Med experts & SPECMOD & 95.00\% & 95.00\% & 85.00\% & 58.75\% \\ 
		\hline
		 All humans & GENMOD & \textbf{99.50\%} & \textbf{98.00\%} & \textbf{96.50\%} & 68.00\% \\ 
		 All humans & SPECMOD & 95.00\% & 95.50\% & 88.00\% & \textbf{68.50\%} \\ 
		\hline
            Consensus Human& GENMOD& \textbf{100.00\%}& \textbf{97.50\%}& \textbf{97.50\%}& \textbf{77.50\%}\\ 
		 Consensus Human & SPECMOD& 95.00\% & 95.00\% & 90.00\% & 72.50\% \\ 
            \hline
		 0-shot Llama2 & GENMOD & \textbf{100.00\%} & 95.00\% & \textbf{92.50\%} & 75.00\% \\ 
		 0-shot Llama2 & SPECMOD & 97.50\% & 95.00\% & 85.00\% & \textbf{80.00\%} \\ 
		\hline
		 1-shot Llama2 & GENMOD & 77.50\% & \textbf{67.50\%} & 47.50\% & \textbf{32.50\%} \\ 
		 1-shot Llama2 & SPECMOD & 77.50\% & 65.00\% & 47.50\% & 22.50\% \\ 
		\hline
		 2-shot Llama2 & GENMOD & \textbf{97.50\%} & \textbf{97.50\%} & \textbf{100.00\%} & 100.00\% \\ 
		 2-shot Llama2 & SPECMOD & 95.00\% & 95.00\% & 97.50\% & 100.00\% \\ 
		\hline
	\end{tabular}
	\caption{Human and LLM Review Scores, reported percentage consistency. The higher the percentage the more often the category was ranked as being consistent.}
	\label{tab:per_consistency}
\end{table*}

\begin{table*}[]
	\centering
	\begin{tabular}{ccccc}
		\hline
		\textbf{Group} & \textbf{FK Age} & \textbf{FK Gender} & \textbf{FK Body Part} & \textbf{FK Coherence}\\ \hline
		 med experts & 0.56 & 0.56 & 0.59 & 0.11 \\ 
		\hline
		 nonmed experts & 1.00 & 1.00 & 0.25 & 0.16 \\ 
		\hline
		 all & 0.72 & 0.72 & 0.41 & 0.18 \\ 
		\hline
	\end{tabular}
	\caption{Fleiss Kappa (FK) among human reviewers}
	\label{tab:fk_human}
\end{table*}

\begin{table*}[]
	\centering
	\begin{tabular}{ccccc}
		\hline
		\textbf{Group} & \textbf{Agr Age}& \textbf{Agr Gender}& \textbf{Agr Lat}& \textbf{Agr Coh}\\ \hline
		 med experts & 97.08\%& 97.08\%& 94.58\%& 72.92\%\\ 
		\hline
		 nonmed experts & 97.50\%& 96.25\%& 88.75\%& 61.25\%\\ 
		\hline
		 all & 97.25\%& 96.75\%& 92.25\%& 68.25\%\\ 
		\hline
	\end{tabular}
	\caption{Percentage Agreement Humans}
	\label{tab:per_agree_human}
\end{table*}

\begin{table*}[]
	\centering
	\begin{tabular}{cccc}
		\hline
		 \textbf{CK Age} & \textbf{CK Gender} & \textbf{CK body part consistency} & \textbf{CK Coherence} \\ \hline
		0.65 & 0.39 & 0.42 & 0.45\\
		\hline
	\end{tabular}
	\caption{Cohen Kappa between two human Medical Experts with similar interpretations of the Coherence criteria}
	\label{tab:ck_human}
\end{table*}

\begin{table*}[]
	\centering
	\begin{tabular}{ccccc}
		\hline
		\textbf{Type} & \textbf{CK Age} & \textbf{CK Gender} & \textbf{CK Body Part}& \textbf{CK Coherence} \\ \hline
		0-Shot Llama& 0.66 & 0.85 & 0.22 & \textbf{0.38}\\
		1-Shot Llama& 0.16 & 0.14 & 0.11 & 0.18\\
		2-Shot Llama& \textbf{0.79}& \textbf{1.00}& \textbf{0.32}& 0.00\\
            3-Shot Llama& 0.00 & \textbf{1.00} & 0.00 & 0.00\\ \hline
            Single Med Expert& 0.79 & 0.49 & 0.82 & 0.58\\
		\hline
	\end{tabular}
	\caption{Cohen Kappa between consensus human choice and Llama-70B choice as well as a randomly selected medical expert human reviewer )}
	\label{tab:ck_llm}
\end{table*}
Table \ref{tab:per_consistency}  presents the reported percentage consistency of GENMOD and SPECMOD as measured by the 5 human reviewers and Llama2. Instead of reporting the results for each individual human reviewer we combine the results into four categories: Medical Experts (3 reviewers), Non-Medical Experts (2 reviewers), All (5 reviewers), and Consensus. The "consensus" human choice is created by selecting inconsistent/consistent for each criteria in the 80 clinical notes (40 notes from each model) based on the choice that the majority of the 5 human reviewers selected. All human reviewers have a preference to GENMOD for age, gender, and body part consistency.  This supports the conclusion that GENMOD contains fewer contradictions inside itself for these categories and results in a more internally consistent note.

For the category of coherence, medical experts preferred SPECMOD while non-medical experts preferred GENMOD.  A medical expert may be more likely to expect that a piece of content should appear in A\&P that was mentioned in HPI (For example, a missing assessment/plan for a postnasal drip that was mentioned in HPI). However, when debriefing the reviewers, they indicated two differing interpretations of coherence. 2 reviewers (both medical experts) interpreted the requirement to be that a note should be marked as incoherent if content was missing from A\&P that was in HPI. The remaining reviewers were only analyzing whether the A\&P contained contradictions to content that appeared in the HPI. The 3 reviewers only analyzing for contradictions preferred GENMOD, while the 2 reviewers looking for all content to be present in both HPI and A\&P preferred SPECMOD. As mentioned in Section \ref{sec:auto_metrics}, this preference for SPECMOD may also be related to the NRNS setting: the evaluation was performed on clinical notes generated with the lower NRNS of 5, which may have prevented GENMOD from producing the content in the A\&P that was present in the HPI.

Table \ref{tab:per_consistency} also shows that for Llama2 evaluations, SPECMOD is only preferred by the 0-shot Llama2 parameter model for the category of coherence. In the 1-shot in-context prompting configuration, Llama was provided only a single positive example for each category, biasing the model to only mark a note as fulfilling the requirement if it was the exact situation mentioned in the example. For example, the in-context prompt for the age category was an example where the age was only mentioned once, for which the in-context prompt provided an answer of "The age is only mentioned once in the HPI section and this age was never contradicted in the note, therefore the answer is TRUE.". However, this biased the model to only mark a note as consistent for age if the age was only mentioned once in the entire note, which caused the model to diverge from human performance.

In order to understand the level of agreement between reviewers, we present the Fleiss Kappa (FK) Inter-Rater Reliability (IRR) via Table \ref{tab:fk_human} and the Percentage Agreement via Table \ref{tab:per_agree_human}. Although agreement is high for age and gender consistency, the agreement is moderate for body part consistency and minimal for coherence. Because FK takes into account the probablity that reviewers could be agreeing with each other by chance, the FK may be low even when the percentage agreement is high. The low FK for coherence is reasonable based upon the discussion earlier: the coherence criteria was interpreted in two different ways which resulted in low agreement. However, if the Cohen Kappa (CK) is calculated between the 2 medical experts that indicated similar interpretations of the coherence category, Table \ref{tab:ck_human} results in a higher IRR for coherence. The moderate agreement in body part consistency is due to the ambiguity in how to interpret a situation where the name is slightly different even though the meaning is near the same. For example, if the CC contained "Left foot pain" but the HPI referred to "Left big toe pain", some reviewers marked the example as inconsistent while others marked it as consistent.

Although every human reviewer had the same interpretation of how to score the categories of age and gender, the FK shows that their agreement is not perfect. This illustrates the difficulty in human review of specific categories: a manual review of the notes with conflicting human review for age and gender revealed that the choice selected by the majority of reviewers was correct, but at least one reviewer made a mistake and selected the wrong value.

We report the CK measurements of the consensus human choice vs Llama via Table \ref{tab:ck_llm}, as well as the CK of a random human reviewer compared to the consensus human choice.  For the categories of age and gender, a manual review confirmed that the consensus selection was always correct. However, the randomly chosen human reviewer did not always make the correct selection, resulting in a CK Age that is the same as the 2-shot Llama model, and a CK Gender worse than the 2-shot Llama.  

The CK scores of 0 for 2 and 3-shot Llama is due to the model always selecting "consistent" for this category, which the CK statistic ranks as completely uncorrelated because the Llama performance is then no-better than a purely random selection, even if the percent agreement between Llama and the consensus human is moderately high. These results show that Llama2 results are correlated to human preferences.

\section{Conclusion}

The paper presents an empirical study on the quality of two distinct PEGASUS-X based designs for generating a clinical note. The GENMOD design reduces the ROUGE by less than 1\% compared to SPECMOD, and has no difference in Factuality. Based on human review, our findings indicate that GENMOD improves the measured age, gender, and body part consistency by 4.5\%, 2.5\%, and 8.5\%, respectively, when compared to SPECMOD.  We highlight the difficulties of measuring the consistency of topics such as note coherence, as the criteria can be difficult to clearly specify in a way that is reliably understood by humans as well as LLMs. Lastly, we observe a Cohen Kappa inter-rater reliability of 0.79, 1.00, and 0.32 for consistency of age, gender, and body part injury, which shows that Llama2 can be a valuable proxy for human evaluation in specific evaluation criteria. This finding supports the usage of the Llama2 LLM as an evaluator on large datasets that would be impractical for comprehensive human review. This proves a promising area for future work to explore development of a robust evaluation suite that utilizes an LLM for clinical note analysis on an expanded variety of criteria.

\section{Limitations}

SPECMOD and GENMOD are trained using DoPaCo transcripts created by an ASR system. Although this does not impact their results compared to each other (since both SPECMOD and GENMOD utilized the same dataset), their absolute performance may be impacted by the quality of the ASR system, since ASR errors may decrease the generated note quality.

The human evaluation only includes 5 human reviewers of the 40 DoPaCos. Because of this, differences of rating on even a few encounters can have an out-sized impact on the results. The human and LLM review was performed on very specific categories and do not necessarily expand to other topics. The prompts used with the Llama2 LLM were developed using a dataset that contained similar specialties and physicians to that which existed in the 40 DoPaCos from the test set. The performance of Llama2 may change if it is used with clinical notes from physicians or specialties that do not exist in the dataset where the prompts were developed.

\bibliography{anthology,custom}

\appendix

\section{Effect of Llama2 Model Size and Quantization on Quality}
\label{sec:size}
Table \ref{tab:ck_llm_quant} present the Cohen Kappa of the various Llama2 sizes and quantization configurations. The prompts were developed using the Llama 70B 16 bit model and the same prompts were used on all configurations, which may further contribute to the higher performance of the 70B parameter model. 

\begin{table*}[]
	\centering
	\begin{tabular}{cllcccc}
		\hline
		\textbf{Prompting Type}   &\textbf{Model Size}&\textbf{Bits}& \textbf{CK Age} & \textbf{CK Gender} & \textbf{CK BP}& \textbf{CK Coh}\\ \hline
		0-Shot&7&16& 0.00 & 0.05 & 0.01 & 0.07\\
		1-Shot&7&16& 0.00 & 0.16 & -0.06 & 0.16\\
		2-Shot&7&16& 0.00 & 0.00 & 0.00 & 0.00\\
		3-Shot&7&16& 0.00 & 0.25 & -0.02 & 0.07\\
		0-Shot&13&16& 0.00 & 0.49 & 0.04 & 0.00\\
		1-Shot&13&16& 0.49 & 0.48 & -0.02 & 0.25\\
		2-Shot&13&16& 0.00 & 0.00 & 0.00 & 0.00\\
		3-Shot&13&16& 0.00 & 0.00 & 0.00 & 0.00\\
		0-Shot&70&8& 0.00 & 1.00 & 0.22 & 0.19\\
		1-Shot&70&8& 0.42 & 0.24 & 0.12 & \textbf{0.38}\\
		2-Shot&70&8& \textbf{0.79} & 0.49 & \textbf{0.32} & 0.00\\
		3-Shot&70&8& 0.66 & 0.49 & 0.00 & 0.00\\
		\hline
	\end{tabular}
	\caption{Cohen Kappa between consensus human choice and LlaMA2 choice}
	\label{tab:ck_llm_quant}
\end{table*}
Table \ref{tab:ck_llm_quant} shows the performance ranking of other Llama models that were tested as a part of this experiment.

\begin{table*}[]
	\centering
	\begin{tabular}{cllccccc}
		\hline
		\textbf{Group} &   \textbf{Model Size}&\textbf{Bits}&\textbf{Model} & \textbf{Age} & \textbf{Gender} & \textbf{BP}& \textbf{Coh}\\ \hline
		 0-Shot&   70&8&GENMOD & 100.00\% & 97.50\% & 92.50\% & 85.00\% \\ 
		 0-Shot&   70&8&SPECMOD & 100.00\% & 95.00\% & 85.00\% & 97.50\% \\ 
		\hline
		 1-Shot&   70&8&GENMOD & 92.50\% & 72.50\% & 55.00\% & 72.50\% \\ 
		 1-Shot&   70&8&SPECMOD & 90.00\% & 82.50\% & 65.00\% & 75.00\% \\ 
		\hline
		 2-Shot&   70&8&GENMOD & 97.50\% & 100.00\% & 100.00\% & 100.00\% \\ 
		 2-Shot&   70&8&SPECMOD & 95.00\% & 97.50\% & 97.50\% & 100.00\% \\ 
		\hline
		 3-Shot&   70&8&GENMOD & 100.00\% & 100.00\% & 100.00\% & 100.00\% \\
		 3-Shot&   70&8&SPECMOD & 97.50\% & 97.50\% & 100.00\% & 100.00\% \\
		\hline
		 0-Shot&   13&16&GENMOD & 100.00\% & 97.50\% & 40.00\% & 100.00\% \\ 
		 0-Shot&   13&16&SPECMOD & 100.00\% & 100.00\% & 40.00\% & 100.00\% \\ 
		\hline
		 1-Shot&   13&16&GENMOD & 100.00\% & 97.50\% & 97.50\% & 90.00\% \\ 
		 1-Shot&   13&16&SPECMOD & 95.00\% & 90.00\% & 100.00\% & 90.00\% \\ 
		\hline
		 2-Shot&   13&16&GENMOD & 100.00\% & 100.00\% & 100.00\% & 100.00\% \\ 
		 2-Shot&   13&16&SPECMOD & 100.00\% & 100.00\% & 100.00\% & 100.00\% \\ 
		\hline
		 3-Shot&   13&16&GENMOD & 100.00\% & 100.00\% & 100.00\% & 100.00\% \\ 
		 3-Shot&   13&16&SPECMOD & 100.00\% & 100.00\% & 100.00\% & 100.00\% \\ 
		\hline
		 0-Shot&   7&16&GENMOD & 100.00\% & 62.50\% & 27.50\% & 97.50\% \\ 
		 0-Shot&   7&16&SPECMOD & 100.00\% & 60.00\% & 22.50\% & 100.00\% \\ 
		\hline
		 1-Shot&   7&16&GENMOD & 100.00\% & 90.00\% & 97.50\% & 90.00\% \\ 
		 1-Shot&   7&16&SPECMOD & 100.00\% & 92.50\% & 92.50\% & 97.50\% \\ 
		\hline
		 2-Shot&   7&16&GENMOD & 100.00\% & 100.00\% & 100.00\% & 100.00\% \\ 
		 2-Shot&   7&16&SPECMOD & 100.00\% & 100.00\% & 100.00\% & 100.00\% \\ 
		\hline
		 3-Shot&   7&16&GENMOD & 100.00\% & 95.00\% & 97.50\% & 97.50\% \\ 
		 3-Shot&   7&16&SPECMOD & 100.00\% & 95.00\% & 100.00\% & 100.00\% \\ 
		\hline
	\end{tabular}
	\caption{Human and LLM Review Scores, reported percentage consistency. The higher the percentage the more often the category was found to be consistent}
	\label{tab:irr_human_quant}
\end{table*}

\section{System Prompts}
\label{sec:sys_prompts}

The following system prompts are used for each evaluation criteria presented in \ref{sec:eval}

\begin{enumerate}
    \item Age: "You are a medical assistant. You will be given a clinical note and should decide whether the age of the patient remains consistent throughout the note. If consistent, answer TRUE. If the age changes between CC and HPI, answer FALSE. If the age is only mentioned once, that means it is consistent and you should answer TRUE. @@NAME@@ is a de-indentification and does not have impact on the gender. Only pay attention to ages that are explicitly stated, do not infer age from any dates provided. Answer in a single word (TRUE or FALSE):"
    \item Gender: "You are a medical assistant. You will be given a clinical note and should decide whether the gender is consistent throughout the note. If the pronoun or gender of the patient is different between the HPI and A\&P, that means the answer is FALSE. Answer in a single word (TRUE or FALSE):"
    \item Body Part consistency: "You are a medical assistant. You will be given a clinical note and should decide whether the injury body part mentioned is consistent throughout the note. For instance, if the CC mentions left foot pain, the HPI should also mention left foot pain. If the visit makes no mention of an injury for which a body part is relevant, then this item is not applicable and you should select TRUE. Answer in a single word (TRUE or FALSE):"
    \item Coherence: "You are a medical assistant. You will be given a clinical note and should decide whether the A\&P section appears to be reasonable based on the HPI section. If the A\&P is reasonable based on the CC and HPI, answer TRUE. Otherwise, answer FALSE. If a condition is mentioned in the CC or HPI that was never addressed in the A\&P Section, answer FALSE. There can be some additional content in the A\&P but it should not be contradictory to the HPI. Answer in a single word (TRUE or FALSE):"
\end{enumerate}
\end{document}